%% file: paper.tex
\documentclass[10pt,twocolumn,letterpaper]{article}
\usepackage{mwe}
\usepackage{cvpr}
\usepackage{times}

\usepackage[utf8]{inputenc} % allow utf-8 input
\usepackage[T1]{fontenc}    % use 8-bit T1 fonts
\usepackage[pagebackref=true,breaklinks=true,letterpaper=true,colorlinks,bookmarks=false]{hyperref}
\usepackage{url}            % simple URL typesetting
\usepackage{booktabs}       % professional-quality tables
\usepackage{amsfonts}       % blackboard math symbols
\usepackage{nicefrac}       % compact symbols for 1/2, etc.
\usepackage{microtype}      % microtypography
\usepackage{amsmath,graphicx}

\usepackage{authblk}
\cvprfinalcopy % *** Uncomment this line for the final submission

\usepackage[utf8]{inputenc} % allow utf-8 input
\usepackage[T1]{fontenc}    % use 8-bit T1 fonts
\usepackage{hyperref}       % hyperlinks
\usepackage{url}            % simple URL typesetting
\usepackage{booktabs}       % professional-quality tables
\usepackage{amsfonts}       % blackboard math symbols
\usepackage{nicefrac}       % compact symbols for 1/2, etc.
\usepackage{microtype}      % microtypography
\usepackage{amsmath,graphicx}
\usepackage{amsmath}
\usepackage{subfig}
\usepackage{color, colortbl}
\usepackage{amsmath}
\definecolor{Gray}{gray}{0.85}
\usepackage{multirow}
\usepackage{authblk}

\newcommand{\expnum}[2]{{#1}\tiny{\mathrm{e}{#2}}}
\newcommand{\name}{MaCC~}
\newcommand{\Xspace}{\mathrm{X}}
\newcommand{\Yspace}{\mathrm{Y}}
\newcommand{\Zspace}{\mathrm{Z}}
\newcommand{\Rspace}{\mathbb{R}}
\newcommand{\F}{\mathcal{F}}
\newcommand{\G}{\mathcal{G}}
\newcommand{\E}{\mathcal{E}}
\newcommand{\D}{\mathcal{D}}
\newcommand{\x}{\mathbf{x}}
\newcommand{\z}{\mathbf{z}}

\title{\textbf{Improved Surrogates in Inertial Confinement Fusion with \\ Manifold and Cycle Consistencies}}

\author{Rushil Anirudh\thanks{Corresponding author: \href{mailto:anirudh1@llnl.gov}{anirudh1@llnl.gov}}, Jayaraman J. Thiagarajan, Peer-Timo Bremer, Brian K. Spears}
 \affil[]{Lawrence Livermore National Laboratory, Livermore, California.}
 \date{}
\begin{document}

\maketitle

\begin{abstract}

Neural networks have become very popular in surrogate modeling because of their ability to characterize arbitrary, high dimensional functions in a data driven fashion. This paper advocates for the training of surrogates that are consistent with the physical manifold -- i.e., predictions are always physically meaningful, and are cyclically consistent -- i.e., when the predictions of the surrogate, when passed through an independently trained inverse model give back the original input parameters. We find that these two consistencies lead to surrogates that are superior in terms of predictive performance, more resilient to sampling artifacts, and tend to be more data efficient. Using Inertial Confinement Fusion (ICF) as a test bed problem, we model a 1D semi-analytic numerical simulator and demonstrate the effectiveness of our approach. Code and data are available at: \href{https://github.com/rushilanirudh/macc/}{https://github.com/rushilanirudh/macc/}
\end{abstract}

\section{Introduction}
\input{intro}

%\section{Related Work}
%\input{related}

\paragraph{Surrogate design for Inertial Confinement Fusion (ICF)}
\input{problem}

\section{Methods}
\input{methods}

%\section{Evaluation Strategies}
%\input{evaluation}

\section{Experimental Settings}
\input{results}

\section{Conclusion}
\input{discussion}

\subsection*{Acknowledgement}
This work was performed under the auspices of the U.S. Department of Energy by Lawrence Livermore National Laboratory under Contract DE-AC52-07NA27344.

\section*{Appendix-I}
\begin{figure*}[!htb]
	\includegraphics[trim={0 0 0 0},clip,height=0.9\textheight]{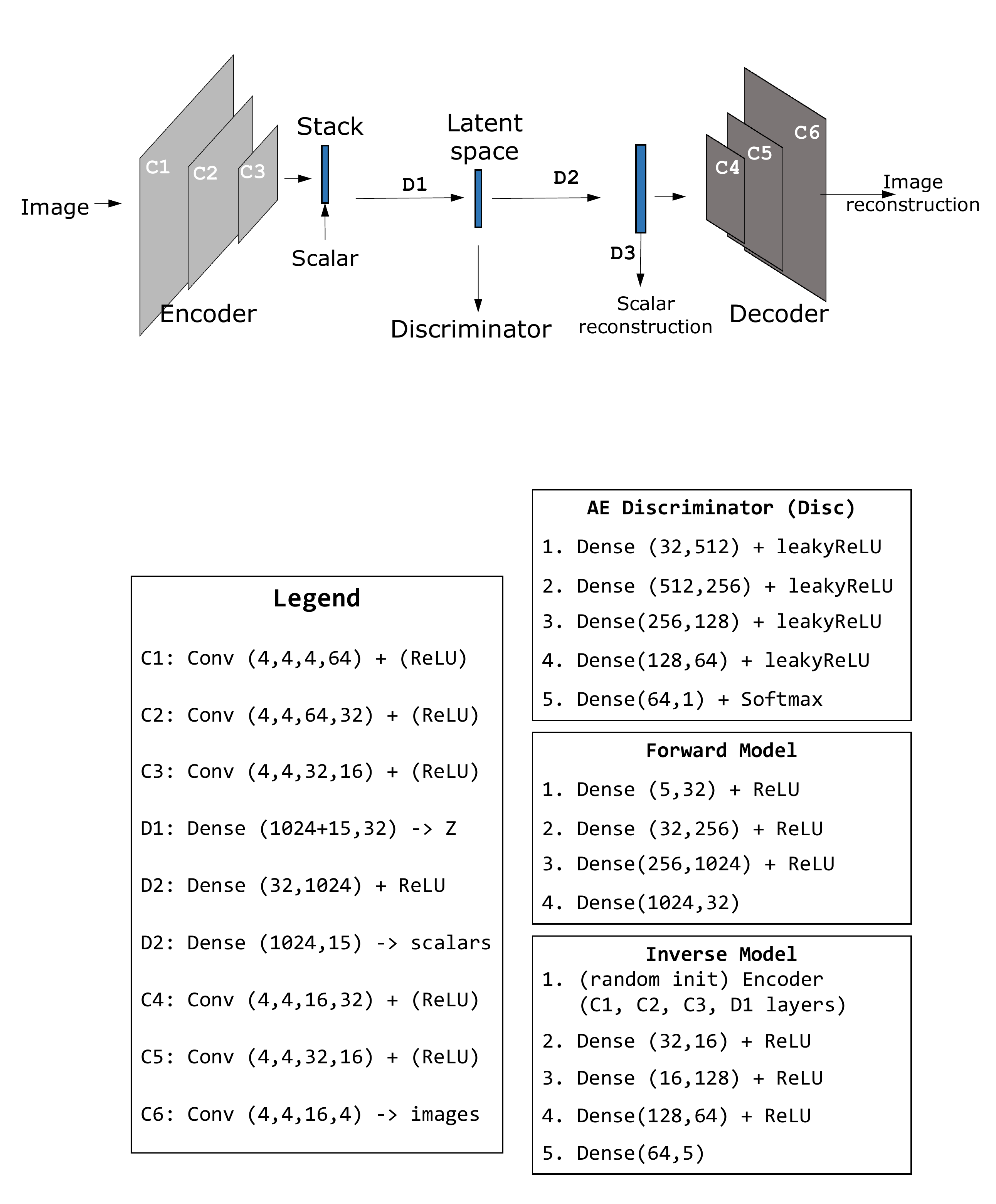}
\caption{Architectural details of all the networks used in our surrogate design.}
\label{fig:architecture}
\end{figure*}
\paragraph{Architecture}
The details of all the networks used here are shown in figure \ref{fig:architecture}.

\section*{Appendix - II} 
\begin{figure*}[!htb]
	\centering
\includegraphics[trim={0 0 0 0},clip,width=\textwidth]{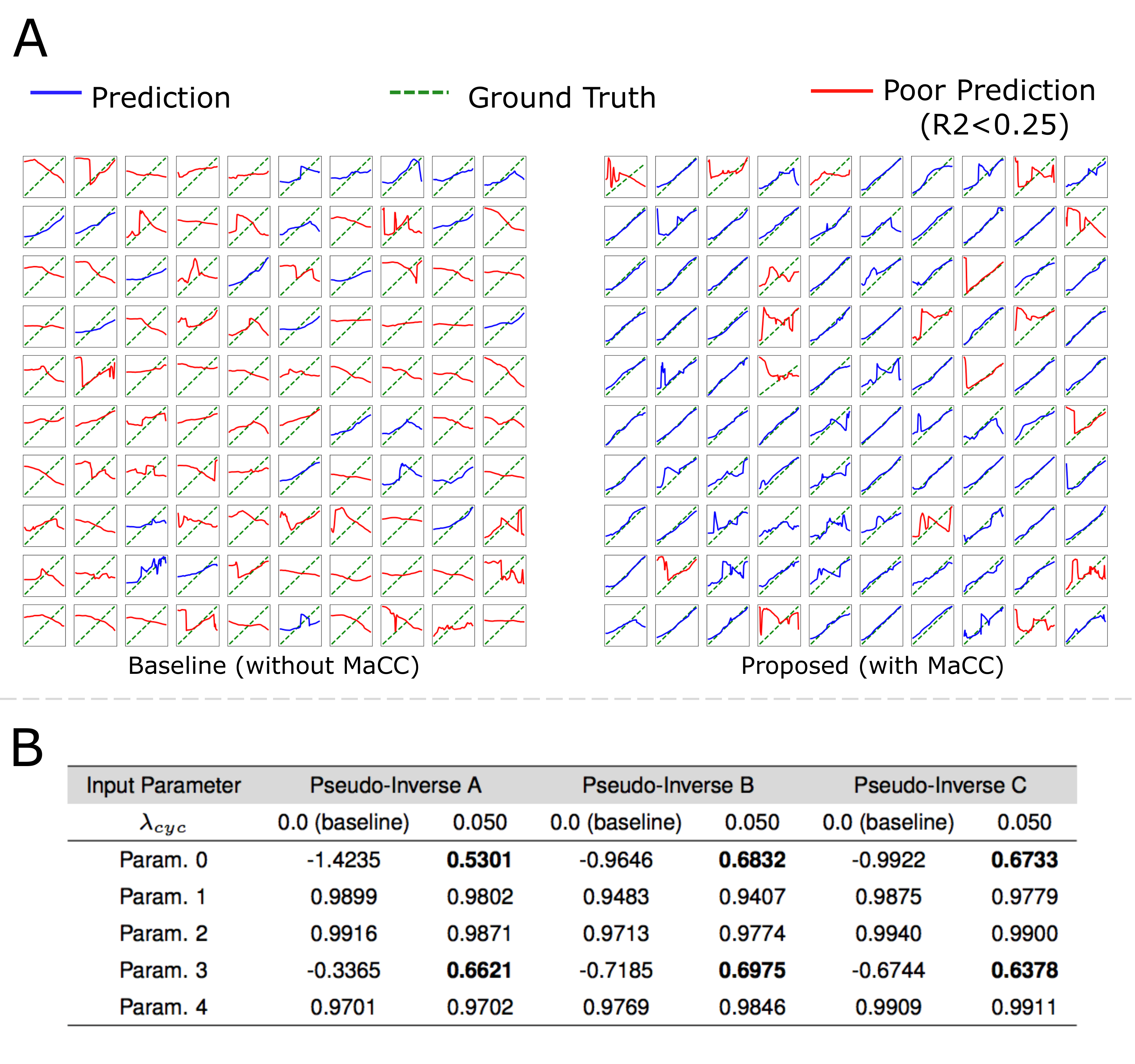}
	\vspace{-5pt}
	\caption{\textbf{A}: Cyclical regularization significantly improves the compatibility between the forward and inverse models, as shown here. run a paramter scan for parameter 3 here for \name and the baseline on a total of 10K samples. Each plot in this figure corresponds to a single sample with 100 variations of parameter 3. A vanilla CNN without cyclical regularization completely fails to capture the trends unlike \name.  \textbf{B:} The results for 1000 random samples across all parameters are shown.}
		\label{fig:cyc_visual}
\end{figure*}

In figure \ref{fig:cyc_visual}-A, we show 100 plots demonstrating the predicted vs true input parameter that is recovered after going through one cycle of the forward, and inverse models. Since we have access to the ground truth, we compare the prediction against it. The ground truth is shown as a dotted line, while the predicted is shown as a solid line. Further, we highlight the poor predictions in red (those which have a prediction accuracy R2-score of less than 0.25). It is evident that the baseline model completely fails this task, with most predictions being poor. It is important to note here that the inverse model is fixed for both the baseline and the proposed models, and they contain the same architecture, and are trained to the same degree of mean squared error on the validation set.

Results averaged across 1000 different samples, for each of the 5 parameters, sampled 100 times (i.e. total of 100,000 artificial samples per dimension) are shown in table \ref{fig:cyc_visual}-B for the three different pseudo-inverse models. We observe that \name~ exhibits self-consistent behaviour across all the models, even though it has never seen them during training.

%%supplementary

\bibliographystyle{ieee}
\bibliography{refs}

\subsection*{Disclaimer}
{\small
 \noindent  This document was prepared as an account of work sponsored by an agency of the United States government. Neither the United States government nor Lawrence Livermore National Security, LLC, nor any of their employees makes any warranty, expressed or implied, or assumes any legal liability or responsibility for the accuracy, completeness, or usefulness of any information, apparatus, product, or process disclosed, or represents that its use would not infringe privately owned rights. Reference herein to any specific commercial product, process, or service by trade name, trademark, manufacturer, or otherwise does not necessarily constitute or imply its endorsement, recommendation, or favoring by the United States government or Lawrence Livermore National Security, LLC. The views and opinions of authors expressed herein do not necessarily state or reflect those of the United States government or Lawrence Livermore National Security, LLC, and shall not be used for advertising or product endorsement purposes. }

\end{document}

%% file: intro.tex
Across scientific disciplines, researchers commonly design and evaluate experiments by comparing empirical observations with simulated predictions from numerical models.
Simulations can provide insights into the underlying phenomena and are often instrumental to effective experiment design. 
Unfortunately, the most reliable, high-fidelity codes are often too expensive to allow extensive calibration or parameter estimation. 
Hence, it is common to use ensembles of simulations to train a surrogate model that approximates the simulator over a large range of inputs, thereby enabling parameter studies as well as sensitivity analysis \cite{paganini2018calogan,peurifoy2018nanophotonic}.
Furthermore, one often fits a second -- {\it inverse} -- model to drive adaptive sampling and to identify parameters that drive the surrogate model into consistency with experiment \cite{peurifoy2018nanophotonic}.  

Until recently, surrogate modeling has largely been restricted to one or at most a handful of scalar outputs. 
Consequently, scientists are forced to distill their rich observational and simulated data into simple summary indicators, or hand-engineered features such as the integral of an image, the peak of a time history, or the width of a spectral line.
Such feature engineering severely limits the effectiveness of the entire analysis chain as most information from both experiments and simulations is either highly compressed or entirely ignored. 
Unsurprisingly, surrogate models designed to predict these features are often under-constrained, ill-conditioned, and not very informative. 

Neural networks (NNs) have become a popular option to address this challenge due to their ability to handle more complex, multi-variate datatypes, such as images, time series, or energy spectra. 
In a number of different application areas ranging from particle physics~\cite{paganini2018calogan} and nanophotonic particle design~\cite{peurifoy2018nanophotonic} to porous media flows~\cite{zhu2018bayesian} or storm predictions~\cite{kim2015time} NNs are able to effectively capture correlations across high dimensional data signatures and produce high quality surrogates, predictors, or classifiers.
% PTB: not sure what this is supposed to tell us? 
%Additionally, NNs are differentiable by design making it easier to add into existing machine learning workflows. 
As a result there has been renewed interest in building better surrogates for scientific problems.
These include incorporating known scientific constraints into the training process~\cite{zhu2019physics,raissi2019physics}, or reducing dimensionality for better uncertainty quantification~\cite{tripathy2018deep}. 
However, over-parameterized NNs are known to require large amounts of training data, and in high dimensional problems, can produce brittle models, wherein perturbations to the data can lead to completely unexpected results, e.g., \textit{adversarial} corruptions~\cite{goodfellow2014explaining}.
Furthermore, the surrogate forward models are often inconsistent with the inverse, leading to an implausible overall system in which the intuitive cycle of mapping inputs to outputs and back to inputs produces wildly varying results. Not only can an inverse prediction from the surrogate output be far away from the initial input, but even univariate sensitivities, i.e. inferring changes in predictions with respect to a single input parameter, are often unintuitive. 
Due to these shortcomings, many scientists are reluctant to adopt NNs surrogates despite their undeniable power in modeling complex relationships.

This paper advocates for the training of Manifold \& Cyclically Consistent (MaCC) surrogates using a multi-modal and self-consistent neural network that significantly outperforms the current state-of-the-art on a wide range of metrics. 
We find that \name surrogates are not only more resilient to sampling artifacts, but also have better predictive performance even in small sample regimes. 
\name surrogates contain two distinct components: (a) An autoencoder to approximate the low-dimensional latent manifold, which accurately captures the correlations between multi-modal outputs of a simulator, i.e.\ multiple images and a set of scalar quantities; and (b) An inverse (or pseudo-inverse) neural network that trains alongside the surrogate network. The direct coupling between forward and inverse model enables us to enforce cyclical consistency, which regularizes the training to produce higher fidelity and more robust models. 
Using a semi-analytic model of inertial confinement fusion~\cite{gaffney2014thermodynamic,kritcher2014metrics} as a test bed problem, we show that our approach produces high-fidelity surrogates, both in terms of standard error metrics, i.e.\ an $L_2$-norm, as well as in terms of self-consistency and stability.

%%% Local Variables:
%%% mode: latex
%%% TeX-master: "main"
%%% End:

%% file: problem.tex
\begin{figure*}[!htb]
\centering
\includegraphics[trim={1cm 1cm 0.5cm 1cm},clip,width=0.95\linewidth]{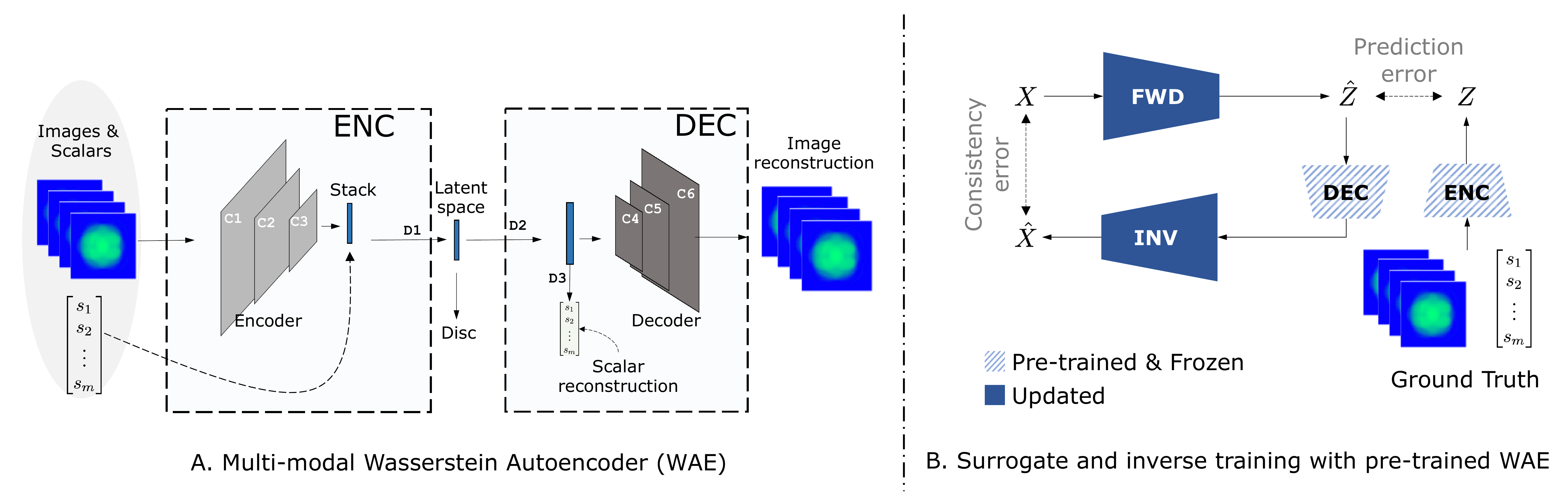}
\caption{\name Surrogates: Proposed architecture uses a pre-trained autoencoder (\textbf{A}) for ensuring manifold consistency, and an inverse model for cyclical consistency and robustness (\textbf{B}).}
\label{fig:overall}
\vspace{-10pt}
\end{figure*}

In any surrogate-based technique, the challenge is to build a high-fidelity mapping from the process inputs, say target and laser settings for ICF, to process outputs, such as ICF implosion neutron yield and X-ray diagnostics.  
Developing surrogates in the ICF context is particularly challenging.
The physics of ICF fusion ignition is predicated on interactions between multiple strongly nonlinear physics mechanisms that have multivariate dependence on a large number of controllable parameters.  
This presents the designer with a complicated response function that has sharp, nonlinear features in a high-dimensional input space.
While this is challenging, deep neural network solutions have made building surrogates for scalar-valued outputs relatively routine~\cite{humbird2018deep}. However, to take full advantage of the rich range of diagnostic techniques, we require surrogates that can also replicate a wide range of array-valued image data. In ICF, the images can be produced by different particles (x-rays, neutrons) at different energies (hyperspectral), at different times and from different lines of sight.  
These complicated modalities are more difficult to ingest, and techniques for learning them can introduce large model capacity and an associated need for excessive amounts of data. 
Thus, our principal design task is to develop a neural network surrogate that can handle multiple data modalities, can produce predictions acceptable for precision physics, and that can be trained without requiring unreasonably large amounts of data.

%%The surrogate modeling problem is being able to design an accuate, reliable, and consistent predictive model that is able to replicate the performance of the scientific phenomenon of interest. Since obtaining large amounts of real scientific data is prohibitively expensive, we use computational simulators in their place. {\color{blue} {\color{blue}
%References for JAG: 1: Betti et al., Physics of Plasmas 9, 2277 (2002)
%2: Springer et al. EPJ Web. Conferences 59:04001 (2013)
%3: Betti et al. Physical Review Letters 114:255003 (2015)
%4: Ott etc al. Physical Review letters 29:1429 (1995)
%}
\paragraph{Predictive Surrogates with Neural Networks} Formally, the surrogate modeling problem is defined as follows: Given a set of input parameters, $X \subset \mathcal{X}$ (obtained with an experiment design of choice, e.g. latin hypercube sample), and the corresponding observations or outputs from the simulator, $Y\subset \mathcal{Y}$, where $Y$ denotes a collection of images $(Y_{img})$ and scalar quantities $(Y_{sca})$, the task is to determine a function $\mathcal{F}: X \mapsto Y$, such that a user-defined measure of predictive accuracy, i.e.\ mean squared error (MSE), is minimized. Here, $\mathcal{X}$ and $\mathcal{Y}$ refer to the space of inputs and outputs respectively.
We refer to $\mathcal{F}$ as the forward model, and the reverse process, $\mathcal{G}: Y \mapsto X$ as the inverse model. 
In many scientific problems a functional inverse may not exist because of the ill-posed nature of the problem, and in such cases we refer to $\G$ as a \emph{pseudo-inverse}.
In recent years, deep neural networks have emerged as the most powerful predictive modeling tool because of their ability to approximate non-linear, and high dimensional functions.
Neural networks are modeled as a series of weights and non-linearities that take the input parameters while predicting the outputs.
They are most commonly optimized using stochastic gradient descent (SGD) with a loss function such as the mean squared error.
However, since neural networks tend to be heavily parameterized, and trained in a stochastic manner, one can expect large variations in the quality of the resulting model as measured by standard performance metrics. 
%Neural networks also tend to be highly opaque in that often their predictions are not easily explainable.

To address this challenge, we propose two novel consistency requirements for predictive surrogate modeling.
First, a manifold consistency that ensures the predictions are physically meaningful; and second, a notion of cyclical consistency between the forward and inverse models.
For the former, we use an autoencoder to embed all output quantities into a low dimensional manifold, $\mathcal{Z}$, and repose surrogate modeling as $\mathcal{F}: X \mapsto Z$, i.e.\ to predict into the latent space rather than directly into $Y$.
To enforce the cycle consistency, we propose to penalize predictions of forward model that are ``inconsistent'' with the inverse model. In other words, a prediction from the forward model, when put through the inverse $\G$ must give back the initial set of parameters, i.e., $\G(\F(X)) \approx X$. These are illustrated in figure \ref{fig:overall}, and described in detail in the next section. 

\paragraph{Notations} Since we have several networks interacting with each other, we clarify our notation for the rest of the paper. We refer to the set of inputs to a simulator by matrix $X$, while each sample is denoted as $\mathbf{x}$. Similarly, the collections of outputs and latent representations are denoted as $Y$ and $Z$, while their individual realizations are $\mathbf{y}$ and $\mathbf{z}$ respectively. The predictions from the trained models $\F$ and $\G$ are referred to as $\hat{\mathbf{y}}$ and  $\hat{\mathbf{x}}$. Finally, we denote a cyclical prediction, i.e. $ \mathbf{x} \rightarrow \hat{\mathbf{y}} \rightarrow \hat{\hat{\mathbf{x}}}$) with a double-hat indicating predictions from both the forward and inverse.

%%% Local Variables:
%%% mode: latex
%%% TeX-master: "main"
%%% End:

%% file: methods.tex
%We present the details of our architecture next. In the following we refer to the forward model as the network that predicts all the output signatures given just the input parameters. The inverse model is the network that takes all the data to predict the likely parameter setting that produced it.

%As discussed above, the goal is to create high quality surrogate models for ICF simulations that incorporate not only scalar summaries, such as {\it energy yield} , but the full set of multi-modal diagnostics, including multiple images and scalars. 
%In this section, we introduce the \name surrogate which improves new neural network based predictive modeling to make it a reliable, and robust surrogate for the ICF simulator.  Rather then directly building a model from $\Xspace$ to $\Yspace$, \name surrogates use an autoencoder to capture correlations between outputs in a reduced dimensional latent space $\Zspace$ and then re-formulates the prediction task as a map from $\Xspace$ to $\Zspace$. We find that this not only makes the learning task easier, but also results in a better surrogate prediction, as measured by mean squared error. Additionally, we propose a new regularization strategy based on enforcing {\it self-consistency} between the forward model $\F: \Xspace \rightarrow \Zspace$ and a pseudo-inverse $\G: \Zspace \rightarrow \Xspace$. Combined, both strategies lead to a more accurate, robust, and a self consistent surrogate.  Next we describe the autoencoder setup, followed by the notion of a pseudo-inverse and how self-consistency leads to more robust predictions. 

\subsection{Multi-modal prediction using an autoencoder}
\label{sec:autoencoder}

Traditional surrogate modeling techniques have largely focused on scalar response functions, $\F: X \rightarrow \Rspace$. 
However, the ultimate goal of a surrogate model is typically to compare results to higher fidelity, i.e.\ experimental data, which will contain a much richer set of diagnostics.
Incorporating the complete set of diagnostics will provide more information and help in better model calibration. 
Furthermore, taking the correlation between multi-modal outputs into account is also expected to lead to a better forward model by disambiguating simulations that may otherwise appear similar in some aggregated response function.

We leverage the recent advances in deep learning to directly build a multi-modal forward model $\F: X \rightarrow Y$. Such a model implicitly has access to the correlation structure present in $\mathcal{Y}$ and will create joint predictions for the different modalities.
However, the task of inferring the correlations from training data is combined with learning the forward model.
Instead, \name splits both tasks by first designing an \textit{autoencoding} neural network to capture the correlation and then explicitly utilize this information to the forward model by predicting into the inferred latent space.
More specifically, we jointly infer an encoder $\E: \Yspace \mapsto \Zspace$ to map a multi-modal observation onto the latent vector $\z \in \Zspace$, and a decoder $\D: \Zspace \mapsto \Yspace$ that reconstructs the multi-modal outputs from the latent representation. %(see Fig.~\ref{fig:autoencoder}).

\paragraph{Design.}
As shown in Fig.~\ref{fig:overall}\textbf{A}, the output space in our setup is comprised of a set of images (treated as different channels) and diagnostic scalars ($s_1, \cdots s_m$).
The encoder is split into two branches: one that uses a convolution neural network to encode image features and another with fully connected layers to process the set of scalars. Both branches are then merged (concatenation) using another fully connected layer to capture the relationships between image features and scalars. The joint output layer forms the overall latent representation and serves as a dimension reduced description of the output space. 
The decoder is built symmetrically to reconstruct the original outputs back. 
In addition to aiming for a high-fidelity reconstruction at the decoder, we encourage the latent space to be approximately uniform by placing a statistical prior in the latent space.  This form of the autoencoder known as a Wasserstein Autoencoder (WAE)~\cite{tolstikhin2017wasserstein}, reduces statistical dependencies between latent factors and helps to regularize the autoencoder training. It also enables us to sample from the latent space efficiently after training.
Mathematically, this is achieved by placing a uniform prior $p(\mathbf{z})$ in the latent space and ensuring that the discrepancy $\mathcal{H}(p(\z), q(\z|\x))$ is minimized, where $\mathcal{H}$ denotes a suitable divergence measure.

Since the exact parameterization of $q(\z|\x)$ is unknown, we adopt an adversarial training strategy (two-sample test), that uses an additional \textit{discriminator} network to ensure that one cannot distinguish between the generated latent representations and realizations from an uniform distribution. 
Formally, the training objective $\mathcal{L}_{ae}$ can be written as:

\begin{gather}
\label{eq:ae}
\sum_{\mathbf{y} \in Y}||\mathbf{y}_{img} - \hat{\mathbf{y}}_{img}||_2^2 +  \gamma_{s}||\mathbf{y}_{sca} - \hat{\mathbf{y}}_{sca}||_2^2 +  \gamma_{a} \mathcal{L}_{adv}, \\
\nonumber \text{where }\z = \mathcal{E}(\mathbf{y}_{img},\mathbf{y}_{sca}),\text{and }\hat{\mathbf{y}}_{img},\hat{\mathbf{y}}_{sca} =  \mathcal{D}\left(\z\right),
\end{gather}and $\mathcal{L}_{adv}$ is the discriminator cross-entropy loss that attempts to classify the latent representation as arising from a \textit{fake} distribution, while assuming the \textit{real} distribution to be uniform random \cite{tolstikhin2017wasserstein}. 
$\gamma_{s}$ is a weight chosen to adjust the bias towards images, we fix it at $\gamma_{s} = \expnum{1}{2}$, and $\gamma_{a} = \expnum{1}{-3}$. 
Given a pre-trained autoencoder, we encode all training data to form $(\mathbf{x},\mathbf{z})$ pairs and reformulate the surrogate as learning $\F:X \mapsto Z$.
%As shown in Section~\ref{sec:results} the resulting model significantly outperforms the baseline mapping directly from $\Xspace$ to $\Yspace$.

\subsection{Cyclical Regularization in Surrogates}
\label{sec:cyclical}
While the surrogate model introduced above performs well it is important to recognize a number of implicit assumptions in the process and consider how they might affect the quality of the model. One of the most important and often disregarded assumptions is the choice of loss function used to construct $\F$. In our formulation, the training objective for the surrogate can be expressed as follows:
\begin{equation}
\min_{\mathcal{F}} \rho\left(\F(\x; \theta) - \z\right),
\label{eq:forward}
\end{equation}where $\rho$ denotes a measure of fidelity and $\F$ represents the parameterized surrogate model with parameters $\theta$. 
Partially for convenience and partially due to a lack of prior knowledge on the residual structure, $\rho$ is often chosen to be an $\ell_p$ norm.
This implicitly assumes that the data manifold, i.e. the space of all outputs $\F(\x)$ for $\x \in \Xspace$, is Euclidean which is most certainly not the case. 
%Note that the baseline model and to some extent even the autoencoder have the same challenge since the data manifold in $\Yspace$ is also not Euclidian. 
Furthermore, the choice of norm also assumes a distribution of discrepancies between the model and the ground truth. Specifically, if we express $\F(\x) = \F^*(\x) + \epsilon(\x)$, where $\F^*$ is the ground truth mapping, then choosing, for example, the $\ell_2$-norm is implicitly assuming that $\epsilon$ follows a Gaussian distribution.
In practice, neither the Euclidian space nor the Gaussian error assumptions are likely to be correct.
However, designing a more appropriate and robust loss function in the latent space is difficult especially for the complex, multi-modal data of interest here. 
Accordingly, we propose a new regularization strategy based on {\it self-consistency} to produce more generalizable and robust forward models.

Conceptually, the challenge in using \eqref{eq:forward} to define $\F$ is two-fold: 
First, since we cannot build a customized $\rho$ and the space of $\theta$'s is large there likely exist many different $\F_i$s with an acceptable error that may represent physically better surrogates than the chosen $\F$. 
Second, the true error is unlikely to be isotropic, meaning some deviations from $\F^*$ are more plausible or less damaging than others. 
To choose among these $\F^s$ we impose a cycle consistency requirement defined as follows:
We train a pseudo-inverse of $\F^*$, i.e., $\G: Y \mapsto X$, and introduce a new regularization term $\delta(\F,\G)$ computed as:
\begin{equation}
\delta(\F,\G) = \sum_{\mathbf{x} \in X, \mathbf{z} \in Z} \|\z - \hat{\hat{\z}}\|_2^2 + \|\x - \hat{\hat{\x}}\|_2^2,
\label{eqn:cons}
\end{equation}
where $\hat{\hat{\z}} = \F(\G(\mathcal{D}(\z)))$ and $\hat{\hat{\x}} = \G(\mathcal{D}(\F(\x)))$, are the cyclical predictions for $\z$ and $\x$ respectively.  Note, different from $\F$, the pseudo-inverse takes the decoded outputs $Y$ instead of $Z$. 
Consequently, the optimization objective for \name surrogates can be expressed as:
\begin{equation}
\min_{\F} \rho\left(\F(\x; \theta) - \z\right) + \lambda_{cyc} \delta(\F,\G).
\label{eqn:cycle}
\end{equation}
Note that in general $\G$ cannot be a true inverse since $\F^*$ might not be bijective. 
In this case constructing $\G$ as a function, i.e.\ a neural network, induces a mode collapse in the estimated posterior $p(\x|\z)$.  
However, even as a pseudo-inverse $\G$ encodes a better local residual structure than $\F$ alone. 
In other words, some errors $\epsilon(\x)$ could be {\it explained} through small changes in $\x$, which will lead to a small $\delta$, while others may be of similar magnitude but in a direction not commensurate with a smooth $\G$, which will lead to a large $\delta$.
For example, consider an error in $\F$ that puts $\hat{\z}$ outside of the data manifold in $\mathcal{Z}$. 
Since $\G$ is trained on the true data manifold, $\hat{\z}$ represents an out-of-distribution sample which is likely to incur a large error in $\hat{\hat{x}}$, thus leading to a large $\delta$. 

In this context, the bi-directional consistency penalty in \eqref{eqn:cons} encourages the surrogate $\F$ to be consistent with the pseudo-inverse in different ways. The first term, is not affected by the mode collapse in the inverse since it is entirely computed in the output space alone. As a result, it encourages the high dimensional output function to be smoothly varying, while the second term constrains the forward model to make predictions closer to the data manifold. 

%The second component makes the loss anisotropic by effectively weighting the forward error $\|\F^*(\x_i) - \F(\x_i)\|_2^2$ with the gradient of $\G$.
Due to the ill-conditioned nature of the inverse, it takes significantly longer to train than the forward. To address this challenge, we pre-train the inverse network \textit{a priori} and then warm start the process and continue training with the cyclical consistency regularization. During surrogate training, the pseudo-inverse is trained with the following loss:
\begin{equation}
\min_{\G} \sum_{\mathbf{z} \in Z}\rho\left(\G(\mathcal{D}(\z); \theta_{I}) - \x \right) + \lambda_{cyc}\|\z - \hat{\hat{\z}}\|_2^2,
\label{eqn:inverse}
\end{equation}where $\theta_{I}$ is the set of parameters of $\G$, and the other terms are the same as in \eqref{eqn:cons}.
Note that optimizing $\F$ according to ~\eqref{eqn:cycle} necessarily biases the model towards a particular pseudo-inverse $\G$.
However, as will be discussed in more detail below, the resulting $\F$ is highly consistent with a diverse set of $\G^{'s}$, different from the one used during training, constructed by bootstrapping the training data. In other words, by including the consistency regularization, the surrogate $\F$ converges to a solution where the resulting residuals are better guided by the characteristics of $\G$. This achieves the same effect as explicitly constructing a specialized loss function $\rho$ to better fit the data characteristics. As we show in our experiments, surrogates obtained using existing neural network solutions are inconsistent with the inverse model and result in non-smooth, non-robust models in practice.

%%% Local Variables:
%%% mode: latex
%%% TeX-master: "main"
%%% End:

%% file: results.tex
%\begin{figure*}[!th]
%	\centering
%	\includegraphics[trim={0 0 0 0},clip,width=0.95\linewidth]{figs/qualitative_16.pdf}
%	\vspace{-5pt}
%	\caption{\textit{Qualitative evaluation of the surrogate obtained using \name} -- Residual images (absolute), with respect to the ground truth, for $16$ examples (only one energy band shown). The intensities of images for both the baseline (\textbf{B}) and \name (\textbf{P}) are normalized to a global scale. Except for a small number of cases (highlighted with red border), \name produces improved quality predictions, when compared to the baseline.}
%	\label{fig:qualitative}
%\end{figure*}

\label{sec:results}
\paragraph{Dataset.}Our training dataset is comprised of input parameter settings and the corresponding outputs from the semi-analytical ICF simulator described in~\cite{gaffney2014thermodynamic}, where each output is a collection of $4$ multi-energy images sized $64\times 64$, and $15$ diagnostic scalar quantities such as yield, ion temperature, pressure, etc. Our dataset was constructed as a random subset ($100$K samples) of a Latin Hypercube experiment design containing $1$ million samples in the $5$-dimensional input parameter space. All model evaluation is carried out using a held-out $10$K validation set, which contains no overlap with the train set. Next, we describe the training strategies adopted for different components of a \name surrogate in our experiments. Note, all models were trained using the Adam Optimizer \cite{kingma2014adam}, with the learning rate set at $\expnum{1}{-4}$, and the mini-batch size fixed at $128$ samples. The architectures for all the models are included in the supplementary material.

\paragraph{Autoencoder.}The first step of \name is to build the autoencoder for concisely encoding the multimodal output space. For all results reported in this section, the dimensionality of the hidden latent space was fixed at $32$. However we did not observe significant performance variability when the number of dimensions was changed. The actual training was carried out until we observed convergence (in terms of MSE) on the $10$K validation set (at $\approx 600$ epochs).

\paragraph{Inverse Model.} In order to achieve cyclical regularization we utilized an inverse model, whose architecture consists of $4$ convolutional layers (for processing the images) and $4$ fully connected layers, and the features from images and scalars are merged using simple concatenation in the penultimate layer of the network. Given the complexity and highly ill-posed nature of the inverse mapping, we first pre-train the inverse model to convergence and then include it for training the surrogate. The pre-training needed $\approx 2500$ epochs for convergence.

\paragraph{Surrogate Design.}The \name surrogate model maps from the input parameter space to the latent representations of the corresponding outputs (from the autoencoder). Given a prediction in the latent space, we utilize the decoder to produce an estimate of the images/scalars for the simulation. Consequently, the model architecture for the surrogate is comprised solely of fully connected layers with non-linear ReLU activations. The self-consistency constraint is imposed by including the pre-trained inverse model into the training process. During training, the strength of the penalty for self-consistency violation in the loss function, given by $\lambda_{cyc}$, is a critical parameter, which we study in detail in our experiments. For comparison, we consider a baseline that represents the current practice in surrogate modeling -- a deep neural network comprised of convolutional and fully connected layers (the total number of learnable parameters and architecture is exactly the same as \name), but leave the autoencoder parts randomly initialized, and set $\lambda_{cyc} = 0.0$.

\section{Results}
\begin{figure*}[!thb]
	\centering
	\includegraphics[trim={0 0 0 0},clip,width=0.95\linewidth]{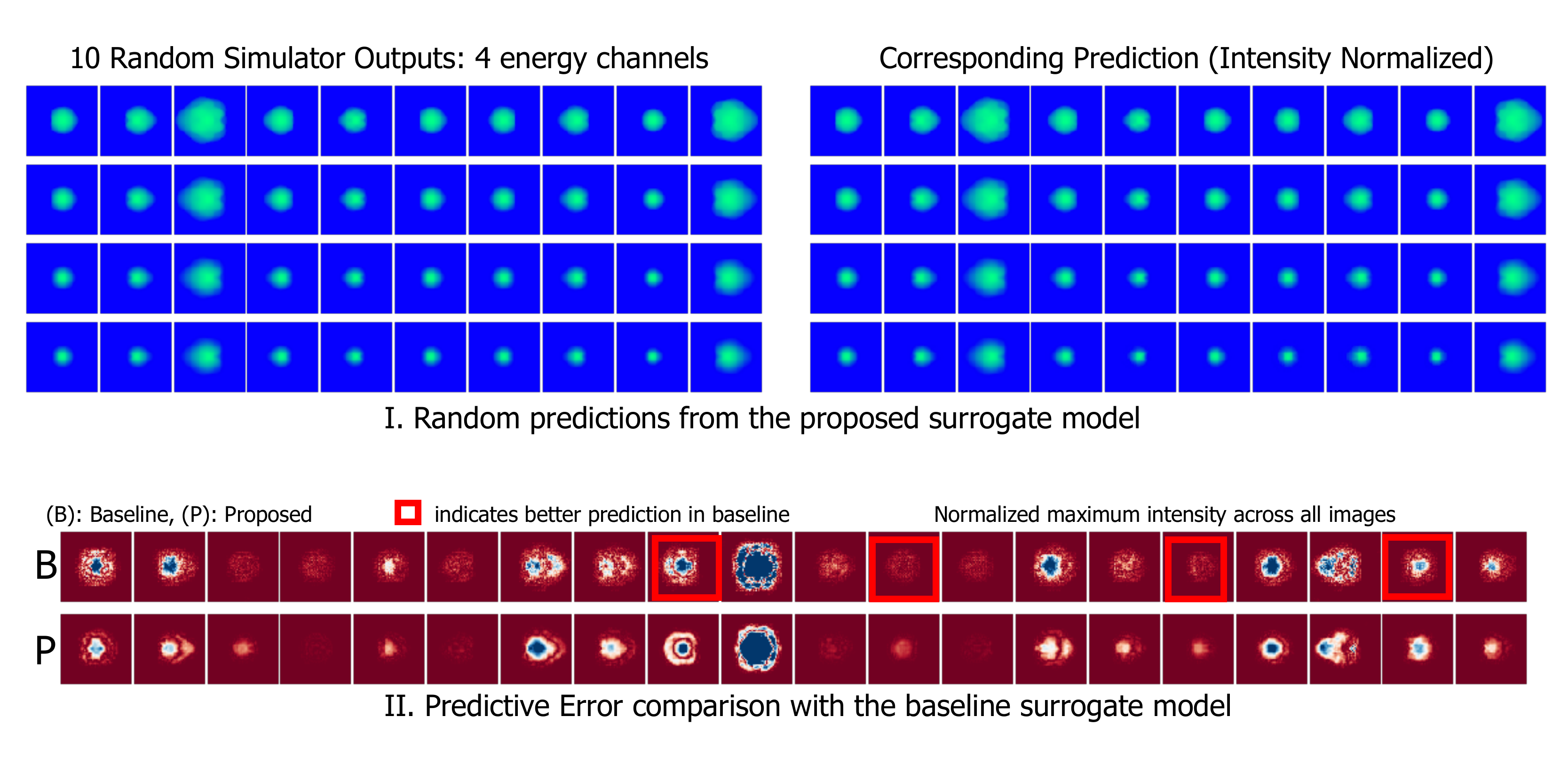}
	\caption{\textbf{I:} The proposed model is able to match the simulator's prediction very closely, across all the four energy bands. Here we show a random sample comparing the simulator's outputs to predictions from a \name surrogate. \textbf{II:}Residual images (absolute), with respect to the ground truth, for $16$ examples (only one energy band shown). The intensities of images for both the baseline (B) and \name (P) are normalized to a global scale. Except for a small number of cases (highlighted with red border), \name produces improved quality predictions, when compared to the baseline.}
	\label{fig:qualitative}
\end{figure*}
\paragraph{Qualitative Evaluation.}Figure \ref{fig:qualitative} (top) shows random samples from the simulator and their corresponding predictions obtained using our surrogate. It can be seen that our surrogate captures all the important details very accurately, across all the energy channels. Next, Figure \ref{fig:qualitative} (bottom) illustrates the residual images for $20$ randomly chosen examples (only one energy band shown) obtained using predictions from the baseline and \name surrogate. All images were intensity normalized by the same maximum intensity value. It is observed that in most cases, \name predicts significantly better quality outputs, where smaller residuals indicate higher fidelity predictions.  

\begin{table*}[!tbh]

	\renewcommand*{\arraystretch}{1.2}
	\begin{minipage}{0.95\linewidth}
		\centering
		\begin{tabular}{c|p{1.5in}|p{1.5in}}
			\hline
			\rowcolor{Gray}
			\textbf{Metric} & \textbf{Baseline (no \name)} & \textbf{Baseline + \name}  \\ \hline
			Mean R2 Scalars & \textbf{0.9990}  & \textbf{0.9974} \\ 
			MSE Image (band 0) & 0.0476 $\pm$ 0.0449 & \textbf{0.0351 $\pm$ 0.0296}  \\
			MSE Image (band 1) & 0.0458 $\pm$ 0.0446 & \textbf{0.0374 $\pm$ 0.0371} \\
			MSE Image (band 2) & 0.08745 $\pm$ 0.1355 & \textbf{0.0736 $\pm$ 0.1236}  \\
			MSE Image (band 3) & 0.2035 $\pm$ 0.4441 & \textbf{0.1742 $\pm$ 0.4010}  \\

			\hline
		\end{tabular}
		\caption{Surrogates with \name show superior predictive performance as measured by mean squared error. Here we use a cyclical weight $\lambda_{cyc} = 0.05$.}
\label{tab:rmse}
\vspace{-15pt}
	\end{minipage}
\end{table*}

\begin{figure}[!htb]
\centering
\includegraphics[trim={0 0 0 0},clip,width=0.80\linewidth]{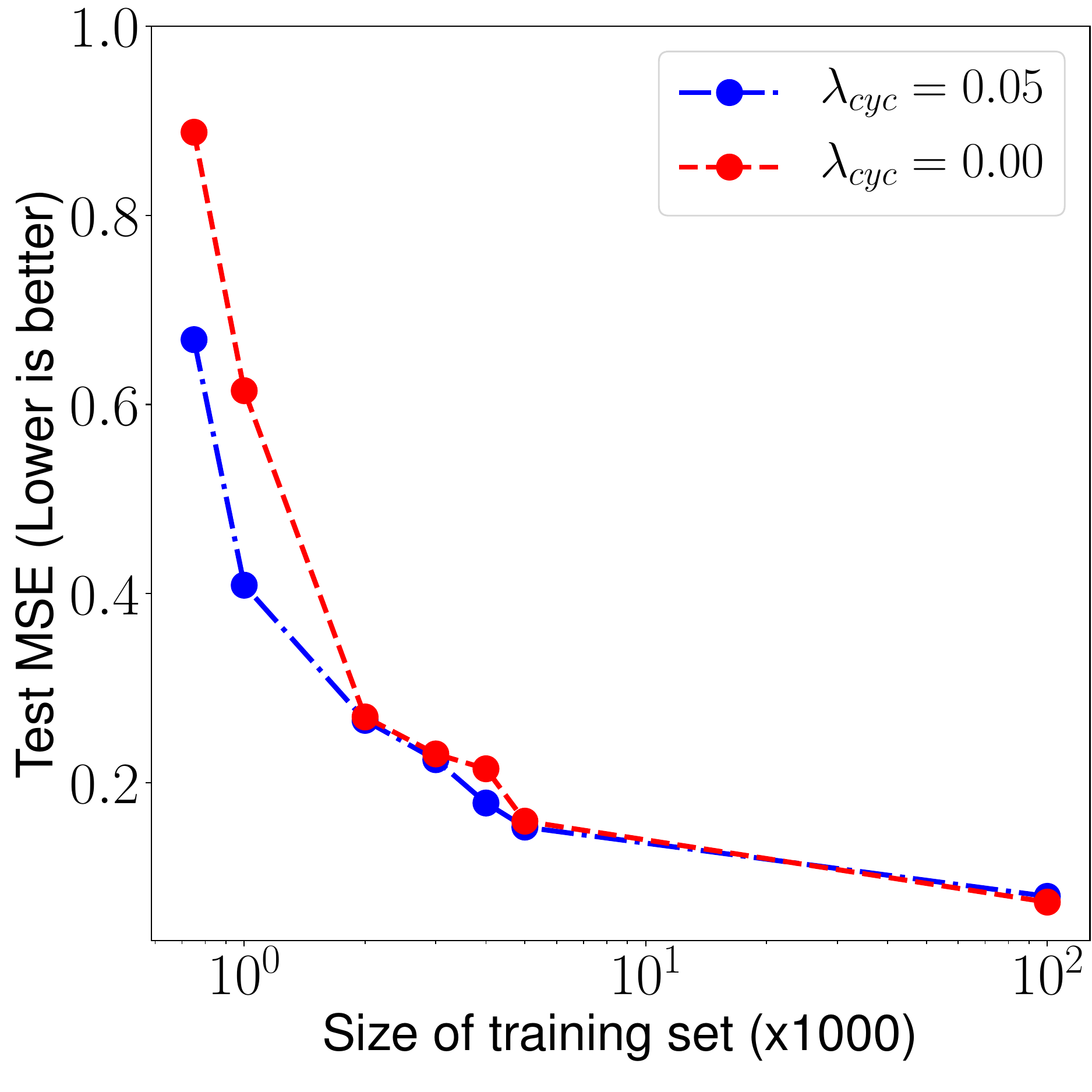}
\caption{Cyclic consistency results in improved generalization with fewer training samples.}
\label{fig:small}	
\vspace{-15pt}
\end{figure}
\begin{figure*}[!htb]
	\centering
\includegraphics[trim={0 0 0 0},clip,width=0.95\linewidth]{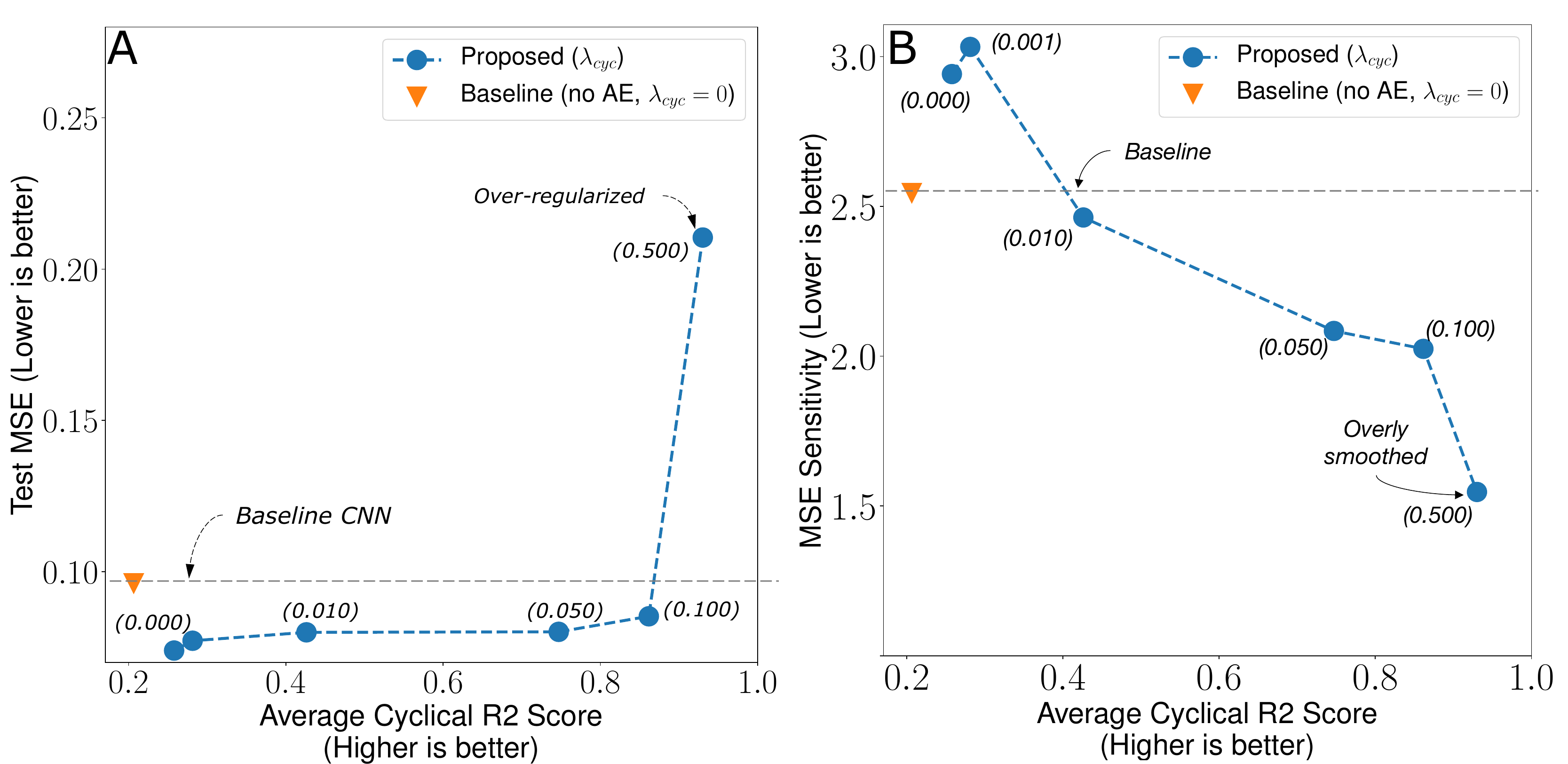}
\caption{\textbf{A.} Ablation study of $\lambda_{cyc}$ and mean squared error. A higher weight leads to more cyclically consistent predictions. Except for extreme cases, the training is fairly robust to values of $\lambda_{cyc}$ leading to a better performance than the baseline. \textbf{B.} Cyclic consistency results in robustness to small local perturbations, as a result of smoothing the high dimensional output prediction space. This also leads to better predictions in smaller data regimes as seen in Figure \ref{fig:small}}
	\label{fig:cyc_plots}
	\centering
\end{figure*}

\paragraph{Quantitative Evaluation.} We evaluate the quantitative performance of the surrogates using widely adopted metrics, namely mean squared error (MSE) and R2. More specifically, we report the following quantities: (a) \textit{Mean R2 Scalars}: Average coefficient of determination (R2 statistic) across the $15$ scalar outputs; (b) \textit{MSE Image (band)}: mean squared error of prediction for the entire $10$K test set, in each of the energy bands. The results are shown in Table \ref{tab:rmse}, where we include the performance of the baseline approach, and \name with $\lambda_{cyc} = 0.05$. From the results for image prediction, it is evident that \name significantly outperforms the baseline neural network solution. In contrast, it is fairly straightforward to predict the scalar diagnostic outputs, with both models achieving an R2 score of $\sim 0.99$. 

\paragraph{Behavior in Small Data Regimes:} Next, we study how cyclical consistency improves generalization properties, while leaving the autoencoder pre-trained, particularly in cases with limited training data. We observe improved predictive performance of the forward model when there are significantly fewer training samples, as shown in Figure \ref{fig:small}. In this experiment, we train different networks while only providing access to a fraction of the training set. It must be noted that the autoencoder is used in this experiment which has been trained on the $100K$ dataset, but it is unsupervised, i.e. it only approximates the physics manifold without any knowledge of the forward process. Finally, we evaluate the performance of this model on the same $10K$ validation set as before in order to make them comparable. Our baseline in this experiment is \name with $\lambda_{cyc} = 0.0$, i.e. without the cyclic consistency. As seen in Figure \ref{fig:small}, cyclic consistency significantly improves the prediction performance in small data regimes -- sometimes by nearly $\sim 30\%$.

\paragraph{A New Self-Consistency Test for Surrogates}
Given the limitations of commonly used error metrics in surrogate evaluation, we introduce a new metric for surrogate fidelity that couples the performance of both the forward and inverse models. Here, we first create a new test set, where we vary only a single parameter using a linear scan, while fixing all other parameters. We linearly sample each dimension in the min and max ranges of that parameter, in 100 steps. These 100 samples are then passed through the forward model, and subsequently through the inverse model before obtaining back input parameter predictions. We then check if the predictions are consistent with the ``ground truth'', which we created. This is conceptually similar to partial dependency tests in statistics and effectively captures sensitivities of the forward and inverse models. 
%This provides a very powerful way to evaluate the quality of the forward and inverse models. 

Given the underdetermined nature of the inverse process, it is possible that the achieved self-consistency is biased by the specific solution of $\G$. Hence, we propose to evaluate the consistency with respect to different solutions from the space of possible pseudo inverse models. To this end, we use multiple random subsets of the original training set (bootstraps) and obtain independent estimates of $\G$. Interestingly, we find that the cyclical consistency remains valid for \name across all of these models, indicating that the self-consistency achieved is actually statistically meaningful. The consistency measure is given by:
\begin{equation}
\label{eq:cyc_score}
\mathcal{L}_c = \sum_{i = 0}^{N_I} \mbox{R2}(\x_{scan}, \G(\mathcal{D}(\F(\x_{scan}))))
\end{equation} 

In our experiments, we used the inverse obtained using $N_I =5$ random samples (each contains a $50\%$ subset) from the training set. Though we show the results for one case, the results for other cases are reported in the supplementary material. In Figure \ref{fig:cyc_plots}-A, we show how cyclical regularization impacts the quality of the surrogate model, against its tendency to be self-consistent. We observe that a small $\lambda_{cyc}$ does not adversely affect the quality of the surrogate model as measured by mean squared error. As it can be seen, until $\lambda_{cyc} = 0.10$ all the models consistently perform better than the baseline. However, with a significantly weight, the model tends to underfit, resulting in a higher MSE.

\paragraph{Robustness via Cyclical Regularization}
We also find that cyclical consistency yields models that are robust to small perturbations in the input parameter space. At test time, we add a small amount of uniform random noise, $\hat{\x} = \x + \sigma*\mathcal{U}$ to the $5$ input parameters, and measure how much the output has changed with regard to the ground truth value at $x$. This is a measure of how smooth the predictions in the output (image) space are. Particularly of relevance to surrogates of scientific models, we expect the function value to change gradually in regions where there are little or no samples around a given test sample. This can be useful in scenarios with sampling artifacts, or a poor design of experiments.  We observe that cyclical consistency has a direct impact on the smoothness of the predictions as shown in Figure \ref{fig:cyc_plots}-B. On the y-axis we show the sensitivity to local perturbations, i.e. the difference in MSE between $\mathcal{F}(\x)$ and $\mathcal{F}(\hat{\x})$, with the consistency measure described in \eqref{eq:cyc_score} on the x-axis. We observe that the cyclical regularization results in significantly more robust models, while having very similar prediction errors on clean data, as seen in Figure \ref{fig:cyc_plots}-A. In order to ensure that the perturbations are not extreme, we pick a $\sigma = 0.1$ for all samples. This was chosen by ensuring that the distance of clean test set to the the perturbed one is smaller than its distance of the nearest neighbor in the training set.

%% file: discussion.tex
In this paper, we introduced \name surrogates, which contain two distinct elements:  a pre-trained autoencoder that enforces the surrogate to map input parameters to the latent space, i.e., $\Xspace\mapsto \Zspace$, instead of the traditional $\Xspace \mapsto \Yspace$; and a pseudo-inverse trained alongside the surrogate with a cyclical consistency objective, which encourages the predictions from $\G(\F(x))$ to be close to the input $x$. These properties lead to robust, data-efficient, and interpretable surrogates, which are properties critical for surrogate models in scientific applications. 